\documentclass[runningheads]{llncs}

\usepackage[T1]{fontenc}
\usepackage[utf8]{inputenc}
\usepackage{graphicx}
\usepackage{booktabs}
\usepackage{array}
\usepackage{multirow}
\usepackage{placeins}
\usepackage{amsmath,amssymb}
\usepackage{url}
\usepackage{xurl}
\usepackage{hyperref}
\begin{document}

\title{Benchmarking LLM-Based Investment Advisor Agents for Asset Allocation in the Chinese ETF Market: Overview of NLPCC 2026 Shared Task 4}
\titlerunning{NLPCC 2026 Shared Task 4: CN-Buzz2Portfolio}

\author{
Liyuan Chen\inst{1,2} \and
Shilong Li\inst{2} \and
Jiangpeng Yan\inst{2}\thanks{Corresponding authors. Email: \{yanjiangpeng, lishilong\}@efunds.com.cn} \and
Shuoling Liu\inst{2} \and
Xi Cheng\inst{2} \and
Haoran Zhang\inst{2} \and
Jianqiu Deng\inst{2} \and
Xiu Li\inst{1} \and
Jian Li\inst{3} \and
Wu Zhu\inst{3} \and
Yingzi Zhu\inst{3} \and
Li Yuan\inst{4} \and
Qianqian Xie\inst{5} \and
Nan Tang \inst{6} \and
Yang Veronica Liu \inst{7} \and
Qiang Yang\inst{7} 
}

\authorrunning{L. Chen et al.}

\institute{
Tsinghua Shenzhen International Graduate School, Shenzhen, China
\and E Fund Management Co., Ltd., Guangzhou, China
\and Tsinghua University, Beijing, China
\and Peking University Shenzhen Graduate School, Shenzhen, China
\and School of Aritificial Intelligence, Wuhan University, Wuhan, China
\and Hong Kong University of Science and Technology (Guangzhou), Guangzhou, China
\and The Hong Kong Polytechnic University, Hong Kong SAR, China
}

\maketitle

\begin{abstract}
Large Language Models (LLMs) convert unstructured financial news into actionable trading strategies, yet standardized end-to-end benchmarks tailored for Chinese market remain scarce. We select common Chinese Exchange-Traded Funds (ETFs) as the investment universe and introduce CN-Buzz2Portfolio, the official benchmark of NLPCC 2026 Shared Task 4. Investment agents take daily trending Chinese financial news and historical ETF price-volume series as inputs to complete news filtering, news-to-ETF mapping and risk-controlled portfolio construction.
To facilitate fair comparison, we provide a reproducible Compression–Perception–Allocation baseline compatible with mainstream LLMs. The task contains two ETF allocation tracks: Macro-Asset Allocation and Sector-Rotation Allocation. We adopt chronological data split across 2024–2026 to alleviate overfitting, with final rankings determined by out-of-sample Sharpe ratios on unseen test data. 26 teams registered for the task, and 14 submitted self-designed agent systems.
Three core findings emerge: (1) Native LLM reasoning capability cannot translate to superior investment returns, as models show highly differentiated risk-return characteristics; (2) Hybrid frameworks are widely adopted, where LLMs extract textual semantic signals and quantitative modules incorporate price-volume factors and risk constraints; (3) Systems based on identical base LLMs yield distinct results with different signal processing, memory and portfolio construction modules. The project is publicly released at \url{https://github.com/splash-li/NLPCC2026-Shared-Task-4}.

\end{abstract}

\section{Introduction}
\label{sec:introduction}

\begin{figure}[t]
\centering
\includegraphics[width=\textwidth]{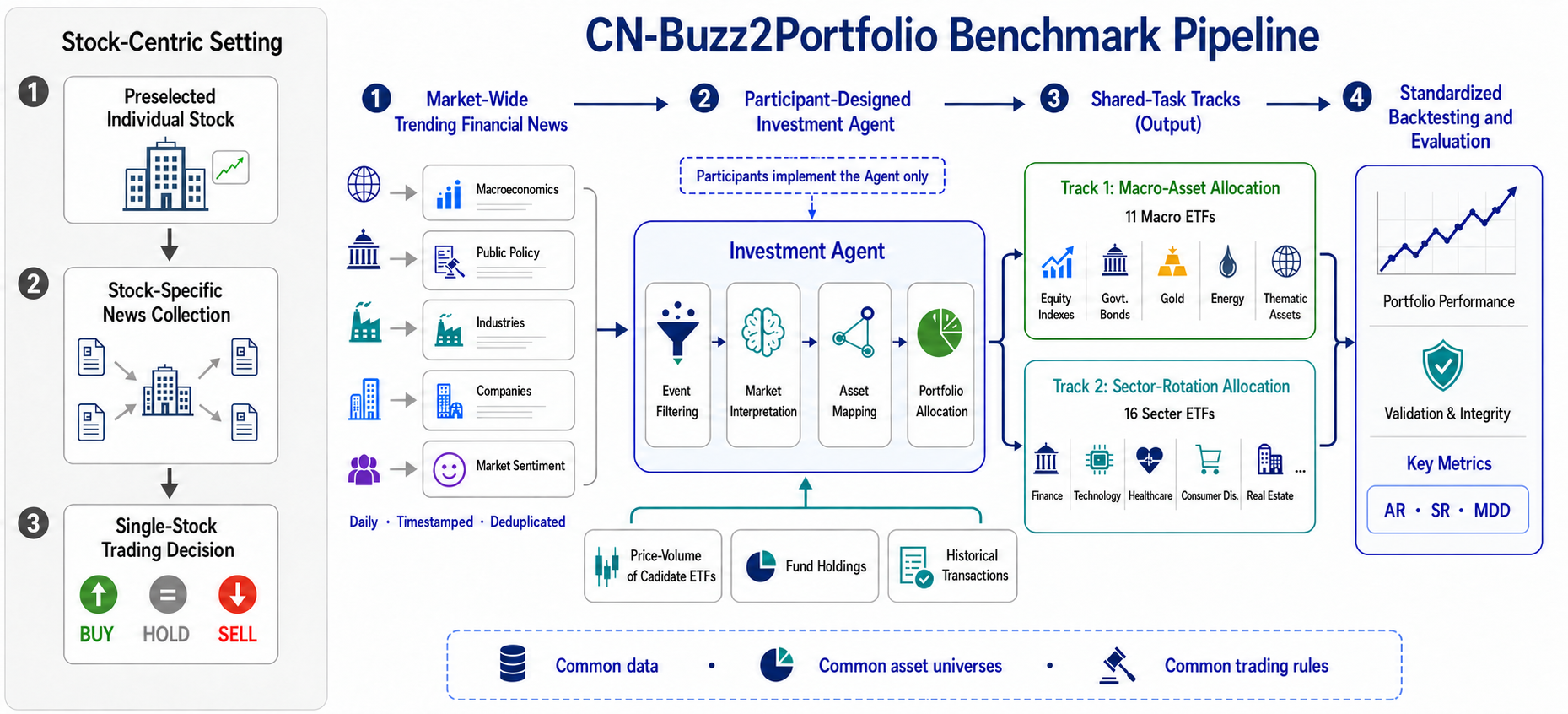}
\caption{Overview of CN-Buzz2Portfolio benchmark pipeline as the NLPCC 2026 Shared Task 4. Two core inputs are provided for investment agents: timestamped market-wide stream of trending Chinese financial news and historical ETF price-volume time series. Participants design Agents that filter events, fuse textual and quantitative market signals, interpret market implications, and construct portfolios for the Macro-Asset Allocation and Sector-Rotation Allocation tracks under a standardized evaluation platform. AR, SR, and MDD denote annualized return, Sharpe ratio and maximum drawdown.}
\label{fig:benchmark-motivation}
\end{figure}

Large Language Models (LLMs) are widely deployed to convert unstructured financial text into actionable portfolio decisions, building on prior research focused on standalone financial sentiment analysis and domain pre-training~\cite{araci2019finbert,wu2023bloomberggpt,yang2023fingpt}. Real-world asset allocation demands a complete sequential decision workflow, including noisy news filtering, cross-market narrative interpretation, asset-narrative matching, portfolio weighting and constrained risk management. In existing studies~\cite{xiao2024tradingagents,li2024deepfund}, investment agents rely on dual sources of input: textual financial news and historical asset price-volume time series to generate reasonable risk-controlled allocation decisions.

Existing benchmarks for LLM trading agents suffer the following major limitations. First, most datasets~\cite{chen2025financialengineering} collect news tied to pre-defined stock candidate pool, and evaluation targets only stock trading. Such designs ignore two core institutional investment scenarios, namely top-down macro asset allocation and industry sector rotation, and fail to test agents’ capability to discover valuable signals from full-market trending information streams. Second, beyond the single-stock bias above, standardized end-to-end benchmarks tailored for Chinese market remain scarce.

To fill this research gap, we build \textit{CN-Buzz2Portfolio}, a standardized benchmark fed with timestamped daily trending news (Chinese market buzz) aggregated from mainstream Chinese financial media platforms. Common Chinese Exchange-Traded Funds (ETFs) are selected as the investment universe for their comprehensive macro/sector exposure and unified trading specifications.
Crucially, the benchmark provides investment agents with two types of parallel inputs: daily trending financial news and historical ETF price-volume time series. This benchmark has been formally adopted as NLPCC 2026 Shared Task 4, titled LLM-Based Investment Advisor Agents for Asset Allocation in the Chinese Market~\cite{nlpcc2026task4}. The shared task sets two complementary tracks, Macro-Asset Allocation and Sector-Rotation Allocation, with independent rankings for each track, and releases a unified trading simulation environment alongside a reproducible organizer baseline framework. A total of 26 teams completed official registration, among which 14 delivered fully functional valid agent submissions.

In particular, the benchmark exposes a unified noisy attention stream to all participants, and requires agents to independently screen useful financial events, fuse text semantics with price-volume quantitative signals, map market narratives to tradable ETF instruments, and output executable portfolio strategies within a shared backtesting environment. Figure~\ref{fig:benchmark-motivation} visualizes the full pipeline linking CN-Buzz2Portfolio dual inputs (news and price-volume series), participant-built agent modules, two allocation tracks and the unified evaluation engine.
To mitigate overfitting to fixed market regimes, the task enforces chronologically separated evaluation. The 2024 dataset (containing paired news and price-volume records) serves for model development, the full-year 2025 public A-board is open for iterative validation, while the 2026 January–June B-board remains unseen by all participants. All submitted agent code is frozen before final assessment, and official rankings are calculated solely by Sharpe ratio produced on the held-out B-board via organizer-side offline evaluation.

This paper summarizes the benchmark and shared task, with the following distinct contributions:
\begin{itemize}
% 修改第一条贡献，补充量价时序数据集
\item \textbf{Benchmark resources.} We construct and release full-market trending Chinese financial news datasets paired with standardized ETF price-volume time series, alongside standardized agent interfaces and an open-source reproducible backtesting platform.
\item \textbf{Shared-task specification.} We design two complementary ETF allocation tracks (Macro-Asset and Sector-Rotation Allocation), unified transaction friction rules and multi-period evaluation protocols for consistent cross-team comparison.
% 修改基线描述，加入量价特征融合
\item \textbf{Reproducible
baseline.} We propose the Compression–Perception–Allocation workflow, which supports joint modeling of LLM text signals and price-volume quantitative features, enabling controlled comparisons across diverse LLMs, classic quantitative strategies and passive market benchmarks.
\item \textbf{Competition overview.} We systematically categorize all submitted agent architectures, report official multi-period rankings, and analyze core empirical patterns around hybrid LLM-quant pipelines that integrate news semantics and price-volume signals, as well as investment agent harnesses.
\item \textbf{Resources.} The full dataset (news and ETF price-volume series), backtesting platform, starter code and all competition documents are publicly released at \url{https://github.com/splash-li/NLPCC2026-Shared-Task-4}.
\end{itemize}
\section{Related Work}

\subsection{Financial Language Models and Decision Agents}

% Financial language models initially emphasized domain representation and text understanding. FinBERT targets financial sentiment analysis~\cite{araci2019finbert}, BloombergGPT studies large-scale financial pretraining~\cite{wu2023bloomberggpt}, and FinGPT develops open financial language models~\cite{yang2023fingpt}. More recent systems extend language models into sequential decision settings. FinMem introduces layered memory for trading agents~\cite{yu2023finmem}; TradingAgents studies specialist-agent collaboration~\cite{xiao2024tradingagents}; and FinRobot provides a general platform for financial applications~\cite{yang2024finrobot}.
Financial language models initially emphasized domain representation and text understanding. FinBERT targets financial sentiment analysis~\cite{araci2019finbert}, BloombergGPT studies large-scale financial pretraining~\cite{wu2023bloomberggpt}, and FinGPT develops open financial language models~\cite{yang2023fingpt}. Guo and Shum formulate the Large Investment Model as a scalable paradigm for learning transferable signals across markets, instruments, and frequencies~\cite{guo2025lim}. More recent systems extend language models into sequential decision settings. FinMem introduces layered memory for trading agents~\cite{yu2023finmem}; TradingAgents studies specialist-agent collaboration~\cite{xiao2024tradingagents}; and FinRobot provides a general platform for financial applications~\cite{yang2024finrobot}. MENTOR decomposes financial event detection, narrative forecasting, and industry-impact prediction through structured multi-agent reasoning~\cite{chen2025mentor}.

\subsection{Financial Decision Benchmarks}

Existing benchmarks cover complementary aspects of financial reasoning and action. AlphaFin organizes stock analysis around retrieved evidence~\cite{li2024alphafin}; InvestorBench evaluates multiple financial decision tasks~\cite{li2024investorbench}; StockBench focuses on historical stock trading~\cite{chen2025stockbench}; and DeepFund evaluates LLMs in fund-investment settings from a live-arena perspective~\cite{li2024deepfund}. CN-Buzz2Portfolio organizes information as a market-wide trending-news stream and evaluates Chinese ETF allocation through a common executable interface in NLPCC 2026 Shared Task 4 as two standardized asset-allocation tracks.

\section{Shared Task Definition and Evaluation Protocol}
\label{sec:competition}
\label{sec:evaluation}

\subsection{Task Formulation}

At each trading date $t$, an agent observes $\mathcal{O}_t=\langle N_t,P_{<t},T_{<t},H_t\rangle$, where $N_t$ is the timestamp-filtered trending-news stream, $P_{<t}$ is historical market data, $T_{<t}$ is the agent's prior trading history, and $H_t$ is its current portfolio. The agent produces target allocations or rebalancing instructions that are validated and executed by the common simulator. The objective is to achieve robust risk-adjusted performance without access to future information.

\subsection{Macro-Asset and Sector-Rotation Allocation Tracks}

The shared task defines two complementary tracks with independent final rankings. Track~1, \textit{Macro-Asset Allocation}, contains 11 broad-market, macro, and thematic instruments and emphasizes macro regimes, policy directions, market styles, and cross-asset defense. Track~2, \textit{Sector-Rotation Allocation}, contains 16 sector ETFs and emphasizes mappings from policy, industry cycles, and supply-chain information to relative sector opportunities. Table~\ref{tab:asset-universes} lists the complete tradable universes.

\begin{table}[t]
\centering
\scriptsize
\setlength{\tabcolsep}{4pt}
\renewcommand{\arraystretch}{1.02}
\begin{tabular}{p{0.12\textwidth}p{0.80\textwidth}}
\toprule
\textbf{Track} & \textbf{Tradable instruments} \\
\midrule
\textbf{Track 1} & CSI 300 (000300.SH); CSI 500 (000905.SH); ChiNext (399006.SZ); STAR 50 (000688.SH); Consumer (000932.SH); New Energy (000941.SH); Media (399971.SZ); Non-ferrous (000819.SH); Energy (000928.SH); Gov Bond (000012.SH); Gold ETF (518880.SH) \\
\midrule
\textbf{Track 2} & Securities (512880.SH); Banks (512800.SH); Insurance (512070.SH); Semiconductor (159995.SZ); AI (159819.SZ); Communication Equipment (515880.SH); Software (159852.SZ); Bio-Pharma (512010.SH); Healthcare (512170.SH); Innovative Drug (159992.SZ); Food \& Beverage (515170.SH); Liquor (512690.SH); Non-ferrous Metals (512400.SH); Coal (515220.SH); Chemicals (159870.SZ); Real Estate (512200.SH) \\
\bottomrule
\end{tabular}
\caption{Tradable instruments for the two shared-task tracks.}
\label{tab:asset-universes}
\end{table}

\subsection{Trading and Evaluation Settings}

All systems start with CNY 100,000 and operate in the same daily-frequency, long-only simulator with a transaction cost of 0.01\%. On trading day $t$, each agent receives only news published before the market close and historical portfolio information, and its orders are executed under the common official-closing-price convention. The platform validates order syntax, available cash, current holdings, and long-only constraints.

The evaluation reports annualized return (AR), Sharpe ratio (SR)~\cite{sharpe1966mutual}, and maximum drawdown (MDD); organizer baseline results additionally report annualized volatility. Metrics are computed from daily portfolio values using 252 trading days per year and a zero risk-free rate. Final rankings are determined independently in the two tracks by B-board Sharpe ratio.

\subsection{Competition Configuration}

The shared task uses the three periods summarized in Table~\ref{tab:competition-timeline}. The 2024 data support sample construction and system configuration, the full-year 2025 A-board provides the public evaluation period, and the B-board covers the hidden final period from 1 January to 1 June 2026.

\begin{table}[t]
\centering
\scriptsize
\begin{tabular}{p{0.21\textwidth}p{0.20\textwidth}p{0.35\textwidth}p{0.14\textwidth}}
\toprule
\textbf{Stage} & \textbf{Period} & \textbf{Participant use} & \textbf{Final rank} \\
\midrule
2024 data & 2024 & Sample construction and configuration & No \\
A-board & 2025 & Public evaluation & No \\
B-board & 2026 Jan.~1--Jun.~1 & Hidden final evaluation & Yes \\
\bottomrule
\end{tabular}
\caption{Competition configuration, applied independently to both tracks.}
\label{tab:competition-timeline}
\end{table}

To prevent implicit knowledge leakage into the B-board, every pretrained model used by an eligible system---including language, embedding, sentiment, and predictive models---must have a documented knowledge cutoff earlier than 1 January 2026. Exact model versions and system configurations are frozen for final evaluation. News is released according to publication timestamps, and decisions may use only task inputs available up to the current simulated trading date.

\section{CN-Buzz2Portfolio Dataset and Evaluation Platform}
\label{sec:dataset-platform}

\subsection{Market-Wide Trending-News Stream}

CN-Buzz2Portfolio aggregates daily high-attention topics from four Chinese financial-information platforms. This attention-oriented design is motivated by evidence that news and investor attention affect trading behavior~\cite{barber2008all}. The stream covers macroeconomic, policy, industry, company, and non-actionable attention signals. Asset mappings are left unspecified, making relevance filtering and asset discovery part of the agent's decision problem.

Each item retains its publication timestamp and is aligned with the simulated trading calendar. Daily duplicate records are normalized under a common preprocessing procedure.

\subsection{Data Statistics}

Table~\ref{tab:data-statistics} reports daily-deduplicated counts and average text lengths for the four sources. The table uses the Top-20 news stream. The organizer CPA baseline uses the Top-10 setting as a fixed input configuration.

\begin{table}[t]
\centering
\small
\renewcommand{\arraystretch}{1.05}
\setlength{\tabcolsep}{5pt}
\begin{tabular}{@{}lrr@{\hspace{18pt}}lrr@{}}
\toprule
\textbf{Source} & \textbf{Count} & \textbf{Avg. Chars.} &
\textbf{Source} & \textbf{Count} & \textbf{Avg. Chars.} \\
\midrule
Caixin & 13,365 & 926.71 & Tencent Stock & 29,284 & 1,720.21 \\
Sina Finance & 35,595 & 1,890.82 & Tiantian Fund & 28,573 & 1,653.94 \\
\bottomrule
\end{tabular}
\caption{Daily-deduplicated news-stream statistics. Counts refer to Top-20 news entries; length is measured in average Chinese characters (Avg. Chars.).}
\label{tab:data-statistics}
\end{table}

\subsection{Standardized Evaluation Platform}

The organizers provide a standardized, ready-to-use evaluation platform covering data access, trade execution, portfolio accounting, and metric calculation. Participants implement the agent that maps each daily observation to target allocations or rebalancing instructions. All systems use the same data, asset universes, and trading rules.

\section{Organizer Baseline System}
\label{sec:organizer-baseline}

\subsection{Compression--Perception--Allocation Workflow}

The organizer baseline holds a three-stage Compression--Perception--Allocation (CPA) workflow fixed while varying the underlying LLM. Its separation of specialist roles follows modular financial-agent designs~\cite{yang2024finrobot,xiao2024tradingagents}. For trading date $t$, the information flow is
\begin{equation}
N_t \xrightarrow{\mathcal{A}_{\mathrm{sum}}} E_t
\xrightarrow{\mathcal{A}_{\mathrm{ana}}} V_t
\xrightarrow{\mathcal{A}_{\mathrm{trade}}} U_t,
\end{equation}
where $N_t$ is the daily news stream, $E_t$ is its compact event representation, $V_t$ contains structured market views, and $U_t$ is a set of executable portfolio instructions.

\paragraph{Compression.} The summarizer $\mathcal{A}_{\mathrm{sum}}$ filters the trending list, removes non-financial or redundant items, and retains a compact set of events with their timestamps and affected themes.

\paragraph{Perception.} The analyst $\mathcal{A}_{\mathrm{ana}}$ interprets the compressed events against the definitions of the candidate assets. It estimates direction, relevance, horizon, and confidence for macro, style, and sector exposures.

\paragraph{Allocation.} The trader $\mathcal{A}_{\mathrm{trade}}$ combines the structured views with historical prices, prior transactions, and current holdings. It produces budget-based buy or ratio-based sell instructions. A deterministic execution layer validates cash, holdings, syntax, and long-only constraints.

CPA provides the organizer's reproducible reference workflow. Participants may use any architecture that complies with the common interface and information restrictions.

\subsection{Baseline Configuration and Results}

The organizer baselines use the common Top-10-news setting. The model set covers standard and reasoning-oriented families including DeepSeek~\cite{deepseek2025v3,deepseek2025r1,deepseek2026v4}, Qwen~\cite{qwen2025qwen3}, GLM~\cite{zai2025glm46}, Gemini~\cite{google2025gemini}, and GPT~\cite{openai2025gpt5}. Table~\ref{tab:baseline-summary} compares the CPA workflow across reasoning-oriented and standard LLMs over all three evaluation periods. The table also reports Momentum~\cite{jegadeesh1993returns}, mean--variance optimization (MVO)~\cite{markowitz1952portfolio}, CSI~300 buy-and-hold, and an equally weighted portfolio.

\begin{table}[!htbp]
\centering
\scriptsize
\setlength{\tabcolsep}{1.7pt}
\renewcommand{\arraystretch}{0.86}
\begin{tabular}{@{}lrrrr|rrrr@{}}
\toprule
& \multicolumn{4}{c|}{\textbf{Track 1: Macro-Asset Allocation}} & \multicolumn{4}{c}{\textbf{Track 2: Sector-Rotation Allocation}} \\
\textbf{Baseline} & Ann. Ret. & SR & MDD & Ann. Vol. & Ann. Ret. & SR & MDD & Ann. Vol. \\
\midrule
\multicolumn{9}{l}{\textbf{2024 data}} \\
\addlinespace[2pt]
\multicolumn{9}{l}{\textit{Reasoning models}} \\
DeepSeek-R1 & 7.82\% & 0.51 & -10.71\% & 17.70\% & 29.84\% & 1.14 & \textbf{-12.93\%} & 25.50\% \\
Qwen3-32B-Think & 5.82\% & 0.50 & -7.66\% & 12.80\% & 22.76\% & 0.94 & -16.21\% & 24.90\% \\
Qwen3-Max-Think & 9.41\% & 0.62 & -10.95\% & 16.70\% & \textbf{42.28\%} & \textbf{1.52} & -15.84\% & 25.10\% \\
DeepSeek-V4-Flash & 6.40\% & 0.54 & \textbf{-7.63\%} & 12.97\% & 7.66\% & 0.41 & -17.29\% & 26.66\% \\
\addlinespace[2pt]
\multicolumn{9}{l}{\textit{Standard models}} \\
DeepSeek-V3 & 9.72\% & 0.62 & -11.48\% & 17.40\% & 7.30\% & 0.42 & -19.51\% & 22.40\% \\
Gemini-2.5-Pro & 11.18\% & 0.67 & -10.89\% & 18.10\% & 19.00\% & 0.80 & -17.10\% & 25.50\% \\
GPT-5 & 7.66\% & 0.53 & -10.79\% & 16.40\% & 25.67\% & 1.07 & -16.35\% & 23.90\% \\
GLM-4.6 & 6.54\% & 0.47 & -12.01\% & 15.90\% & 33.24\% & 1.25 & -14.11\% & 25.40\% \\
Qwen3-32B & 5.85\% & 0.44 & -11.80\% & 15.80\% & 13.91\% & 0.63 & -17.31\% & 25.70\% \\
Qwen3-Max & 8.41\% & 0.60 & -10.31\% & 15.50\% & 27.86\% & 1.03 & -16.02\% & 27.30\% \\
\addlinespace[2pt]
\multicolumn{9}{l}{\textit{Quantitative baselines}} \\
Momentum & 5.72\% & 0.40 & -13.55\% & 14.30\% & 11.90\% & 0.48 & -17.68\% & 25.00\% \\
MVO & \textbf{17.83\%} & \textbf{1.73} & -8.64\% & \textbf{10.30\%} & 18.64\% & 0.87 & -14.39\% & \textbf{21.40\%} \\
\addlinespace[2pt]
\multicolumn{9}{l}{\textit{Market baselines}} \\
CSI 300 B\&H  & 17.00\% & 0.84 & -14.41\% & 21.40\% & 17.00\% & 0.84 & -14.41\% & 21.40\% \\
EW Portfolio B\&H  & 9.65\% & 0.53 & -15.60\% & 21.80\% & 10.27\% & 0.50 & -18.05\% & 26.30\% \\
\midrule
\multicolumn{9}{l}{\textbf{2025 A-board}} \\
\addlinespace[2pt]
\multicolumn{9}{l}{\textit{Reasoning models}} \\
DeepSeek-R1 & 35.56\% & 2.30 & -6.71\% & 13.60\% & 20.54\% & 1.04 & -13.10\% & 19.80\% \\
Qwen3-32B-Think & 36.05\% & \textbf{2.41} & -8.81\% & 13.20\% & 38.48\% & 1.68 & -13.40\% & 20.70\% \\
Qwen3-Max-Think & \textbf{36.26\%} & 2.15 & -7.58\% & 14.90\% & 28.26\% & 1.28 & -15.61\% & 21.20\% \\
DeepSeek-V4-Flash & 34.52\% & 2.36 & -6.31\% & 12.99\% & 21.23\% & 1.12 & -13.16\% & 18.78\% \\
\addlinespace[2pt]
\multicolumn{9}{l}{\textit{Standard models}} \\
DeepSeek-V3 & 34.68\% & 2.01 & -7.87\% & 15.40\% & 26.38\% & 1.23 & -13.86\% & 20.80\% \\
Gemini-2.5-Pro & 32.02\% & 2.21 & -5.54\% & 13.00\% & 27.08\% & 1.44 & -10.01\% & 17.80\% \\
GPT-5 & 28.74\% & 1.99 & -7.56\% & 13.10\% & 33.18\% & 1.60 & -12.23\% & 19.10\% \\
GLM-4.6 & 35.22\% & 2.36 & -6.36\% & 13.10\% & 26.87\% & 1.20 & -14.30\% & 21.90\% \\
Qwen3-32B & 19.60\% & 1.98 & -6.76\% & 9.30\% & 36.99\% & 1.61 & -14.87\% & 20.90\% \\
Qwen3-Max & 35.60\% & 1.94 & -8.95\% & 16.40\% & \textbf{43.51\%} & \textbf{1.71} & -16.06\% & 22.60\% \\
\addlinespace[2pt]
\multicolumn{9}{l}{\textit{Quantitative baselines}} \\
Momentum & 2.17\% & 0.20 & -16.56\% & 23.00\% & 1.86\% & 0.16 & -17.83\% & 24.70\% \\
MVO & 3.94\% & 0.98 & \textbf{-3.56\%} & \textbf{8.60\%} & 4.54\% & 0.67 & \textbf{-9.12\%} & \textbf{14.60\%} \\
\addlinespace[2pt]
\multicolumn{9}{l}{\textit{Market baselines}} \\
CSI 300 B\&H  & 3.16\% & 0.50 & -10.49\% & 15.60\% & 3.16\% & 0.50 & -10.49\% & 15.60\% \\
EW Portfolio B\&H  & 5.39\% & 0.77 & -10.51\% & 15.90\% & 5.36\% & 0.66 & -12.59\% & 19.30\% \\
\midrule
\multicolumn{9}{l}{\textbf{2026 B-board (Jan.~1--Jun.~1)}} \\
\addlinespace[2pt]
\multicolumn{9}{l}{\textit{Reasoning models}} \\
DeepSeek-R1 & 3.45\% & 0.27 & -12.34\% & 21.57\% & \textbf{26.51\%} & \textbf{1.05} & -14.61\% & 25.83\% \\
Qwen3-32B-Think & -5.22\% & -0.20 & -11.37\% & 18.54\% & 6.51\% & 0.39 & -14.81\% & 22.98\% \\
Qwen3-Max-Think & 9.96\% & 0.55 & -12.72\% & 21.69\% & 12.91\% & 0.64 & -15.19\% & 23.35\% \\
DeepSeek-V4-Flash & \textbf{10.86\%} & \textbf{0.73} & -9.83\% & 16.05\% & 14.49\% & 0.69 & -14.31\% & 24.00\% \\
\addlinespace[2pt]
\multicolumn{9}{l}{\textit{Standard models}} \\
DeepSeek-V3 & -9.10\% & -0.59 & -12.83\% & 14.47\% & 8.05\% & 0.46 & -12.84\% & 22.74\% \\
Gemini-2.5-Pro & -14.96\% & -0.88 & -11.43\% & 16.99\% & -2.34\% & -0.03 & -13.68\% & 19.40\% \\
GPT-5 & 10.54\% & 0.65 & -10.13\% & 18.15\% & 1.21\% & 0.16 & -12.66\% & 18.95\% \\
GLM-4.6 & 6.46\% & 0.43 & -10.86\% & 18.56\% & 25.41\% & \textbf{1.05} & -12.75\% & 24.58\% \\
Qwen3-32B & -4.77\% & -0.14 & -16.64\% & 20.34\% & -5.56\% & -0.18 & -16.34\% & 20.44\% \\
Qwen3-Max & 8.20\% & 0.51 & -9.42\% & 19.31\% & -6.73\% & -0.16 & -18.00\% & 24.84\% \\
\addlinespace[2pt]
\multicolumn{9}{l}{\textit{Quantitative baselines}} \\
Momentum & -13.54\% & -0.43 & -12.58\% & 26.09\% & 19.90\% & 0.86 & -12.02\% & 25.04\% \\
MVO & 3.17\% & 0.37 & \textbf{-4.16\%} & \textbf{9.95\%} & 12.24\% & 0.88 & \textbf{-7.35\%} & \textbf{14.40\%} \\
\addlinespace[2pt]
\multicolumn{9}{l}{\textit{Market baselines}} \\
CSI 300 B\&H  & 9.88\% & 0.65 & -6.71\% & 16.87\% & 9.88\% & 0.65 & -6.71\% & 16.87\% \\
EW Portfolio B\&H  & -5.83\% & -0.22 & -9.52\% & 19.24\% & -11.25\% & -0.57 & -10.09\% & 18.18\% \\
\bottomrule
\end{tabular}

\caption{Unified organizer provided triple-agent baseline tested with different LLMs across the 2024 data, 2025 A-board, and 2026 B-board. Ann. Ret., SR, MDD, Ann. Vol. denote annualized returns, Sharpe ratio, maximum drawdown and annualized volatility; B\&H is short for buy and hold, EW is short for equal-weighted. Boldface marks the best value for each metric within each period and track (highest return and SR, lowest volatility, and MDD closest to zero).}
\label{tab:baseline-summary}
\end{table}

The two allocation tasks favor different LLMs. Quantitative and market references remain competitive in several settings. These results support evaluating LLMs within complete allocation systems.

\section{Participating Systems}
\label{sec:participating-systems}

\subsection{Participation Overview}

The shared task attracted 26 registered teams, among which 14 submitted valid systems.
Table~\ref{tab:participants_combined} summarizes participating systems, including tracks, disclosed core models, team members and affiliations.

\begin{table}[t]
\centering
\scriptsize
\setlength{\tabcolsep}{3.2pt}
\renewcommand{\arraystretch}{1.25}
\begin{tabular}{
>{\raggedright\arraybackslash}p{0.22\textwidth}
>{\raggedright\arraybackslash}p{0.20\textwidth}
>{\raggedright\arraybackslash}p{0.52\textwidth}
}
\toprule
\textbf{System (Track(s))} & \textbf{Selected Model(s)} & \textbf{Team Member(s) \& Affiliation(s)} \\
\midrule
Turing  (2)
& Qwen3‑32B; \newline DeepSeek‑R1
& Tianqi Long; Bin Li \newline \textit{National University of Singapore} \\

Outlier (1)
& GPT‑5
& Jiawei Yang; Xuesheng Li; Cheng Li; Xiu Li \newline \textit{Tsinghua University; University of International Business and Economics} \\

Texas Hold'em (1, 2)
& DeepSeek‑V4‑Pro
& Minghao Zhong; Jiayong Li \newline \textit{CITIC Wealth Co., Ltd.} \\

FinArbi (2)
& DeepSeek‑V4‑Flash
& Yitian Liu; Lu Li; Leyi Su; Zhou Huang \newline \textit{Peking University; Wuhan University} \\

FinanceAgent (1, 2)
& DeepSeek‑V4‑Flash
& Bin Huang; Yi Luo; Ziliang Yin; Jiayang Luo; Yuxiang Jia; Yingjie Han; Kunli Zhang;  \newline \textit{Zhengzhou University} \\

AlphaAllocator (1, 2)
& Local BGE/FinBERT
& Yuhe Sui \newline \textit{Nanyang Technological University} \\

AEGIS (1, 2)
& Gemini‑2.5‑Flash
& Bohe Chen \newline \textit{Beijing University of Posts and Telecommunications} \\

Dianjin (1, 2)
& DeepSeek‑V3.2‑Exp
& Boyang Li \newline \textit{Alibaba Cloud Tongyi Dianjin Team} \\

lizheng (1, 2)
& DeepSeek‑V4‑Pro
& Zheng Li \newline \textit{Individual participant} \\

Varina777 (2)
& DeepSeek‑V4‑Pro
& Min Liu; Rui Gao; Huimin Xiong \newline \textit{Jiangsu Banjv Technology Co., Ltd.} \\

AgentWeaver (1, 2)
& DeepSeek‑V4‑Flash
& Tianyu Hao \newline \textit{Sun Yat‑sen University; Peng Cheng Laboratory} \\

Grilled Egg Crepe (1, 2)
& DeepSeek‑V4‑Pro/ \newline Flash; Qwen3.7‑Plus
& Xiaohui Hu; Kaidong Yu \newline \textit{China Telecom Artificial Intelligence Technology (Beijing) Co., Ltd.} \\

Shrike (1, 2)
& Frozen multifactor \newline model (no LLM)
& Ruitao Feng; Xudong Hong \newline \textit{Kealios Technology} \\

Uibequant\_strategy (2)
& Qwen‑Plus
& Weishen Jiao \newline \textit{University of International Business and Economics} \\
\bottomrule
\end{tabular}
\caption{The overview of participating systems.}
\label{tab:participants_combined}
\end{table}

\subsection{Quantitative-Centered Systems}

\paragraph{FinanceAgent:} uses quantitative factors as the primary signal and permits only bounded LLM adjustments. Regime-dependent asset pools, momentum and low-volatility factors, confirmation delays, minimum holding periods, and gradual trading stabilize allocations.

\paragraph{AlphaAllocator:} builds Track~1 portfolios with shrinkage-based risk parity and Track~2 portfolios with interpretable rotation factors. Local BGE/FinBERT models estimate news relevance and sentiment for small constrained tilts.

\paragraph{lizheng:} first constructs a portfolio from a trained quantitative strategy library and asset budgets. DeepSeek-V4-Pro acts as a daily reviewer and changes the proposal only under explicit risk conditions; invalid outputs revert to the quantitative portfolio.

\paragraph{Shrike:} is a no-LLM system, which selects technical factors on 2024 data using IC/ICIR, freezes them before evaluation, and equally weights approximately the top third of assets by composite score.

\subsection{LLM Semantic Signals with Quantitative Allocation}

\paragraph{Outlier:} uses GPT-5 to extract structured macro events and asset rankings. A temporal macro knowledge graph, regime detector, multi-signal scorer, and programmatic allocator determine final weights under holding-period and turnover constraints.

\paragraph{Texas Hold'em:} accumulates LLM-derived ETF sentiment across days and combines it with technical signals. Black--Litterman fusion~\cite{black1992global} produces target views, followed by portfolio optimization, concentration checks, and periodic review.

\paragraph{AEGIS:} converts daily news into a time-decayed event graph and propagates effects for up to three hops. Graph signals are fused with momentum and volatility factors before layered execution controls generate trades.

\paragraph{Varina777:} uses XGBoost~\cite{chen2016xgboost} for sector selection, augmented by news-derived semantic factors and a macro regime gate. The LLM may reduce risk budgets, while a final counterfactual risk agent can veto unstable trades.

\paragraph{Grilled Egg Crepe:} combines frozen quantitative rules with tactical LLM review. Rules generate macro or sector candidate portfolios; the LLM interprets news and selects, retains, or rejects candidates without freely generating unconstrained weights.

\paragraph{Uibequant\_strategy:} rebalances weekly to reduce daily noise and turnover. Qwen-Plus interprets sector narratives, while recent price and volume action confirms whether the market supports the proposed rotation.

\subsection{Multi-Stage and Multi-Agent Systems}

\paragraph{Turing.} Turing uses a three-stage news pipeline followed by DeepSeek-R1 decision making. Macro, sector, and technical scores are passed through deterministic position caps, trailing stops, hysteresis, and defensive regime rules.

\paragraph{FinArbi.} FinArbi separates technical and news analysis, resolves disagreements through a signal-arbitration agent, and incorporates historical reflection before a portfolio agent produces the final sector allocation.

\paragraph{Dianjin.} Dianjin uses NSGA-II~\cite{deb2002fast} offline to select and freeze a strategy specification. Online quantitative signals, news sentiment, and bounded LLM market views adjust the portfolio, with smoothing, caps, and a quantitative fallback.

\paragraph{AgentWeaver.} AgentWeaver uses macro, sentiment, and risk specialists with an aggregation agent for Track~1, but a focused single news agent for Track~2. Technical ranking, allocation constraints, and periodically distilled trading principles guide execution.

\section{Official Results and Analysis}
\label{sec:results-analysis}

\subsection{Official Leaderboards}

Tables~\ref{tab:track1-results} and~\ref{tab:track2-results} report the official results and follow the final order determined by B-board Sharpe ratio. A-board scores are included to compare system behavior across the two evaluation periods. Rankings use the unrounded official scores.

\begin{table}[t]
\centering
\scriptsize
\begin{tabular*}{\textwidth}{@{}rl@{\extracolsep{\fill}}rrr|rrr@{}}
\toprule
& & \multicolumn{3}{c|}{\textbf{2025 A-board}} & \multicolumn{3}{c}{\textbf{2026 B-board (final)}} \\
\textbf{Rank} & \textbf{Track 1 system} & SR & AR\% & MDD\% & SR & AR\% & MDD\% \\
\midrule
1 & AlphaAllocator & 2.31 & 24.42 & -5.57 & \textbf{1.11} & \textbf{13.52} & -6.23 \\
2 & Dianjin & 1.97 & 15.84 & -4.56 & 0.78 & 8.68 & -6.59 \\
3 & Texas Hold'em & 1.46 & 27.40 & -10.80 & 0.73 & 13.17 & -10.93 \\
4 & FinanceAgent & 1.58 & 35.28 & -15.10 & 0.49 & 8.85 & -12.35 \\
5 & Outlier & 2.46 & 17.34 & \textbf{-2.28} & 0.40 & 2.75 & -4.80 \\
6 & Shrike & 1.29 & 22.23 & -8.42 & 0.40 & 6.22 & -11.85 \\
7 & AEGIS & \textbf{3.78} & 42.82 & -3.14 & 0.30 & 3.18 & -7.91 \\
8 & Grilled Egg Crepe & 2.98 & \textbf{65.08} & -7.81 & 0.20 & 2.05 & -14.91 \\
9 & lizheng & 1.87 & 8.62 & -2.57 & -0.38 & -2.75 & \textbf{-4.67} \\
10 & AgentWeaver & 2.75 & 20.49 & -2.93 & -0.51 & -4.69 & -6.84 \\
\bottomrule
\end{tabular*}
\caption{Official Track~1 results, ordered by final B-board Sharpe ratio.}
\label{tab:track1-results}
\end{table}

\begin{table}[t]
\centering
\scriptsize
\begin{tabular*}{\textwidth}{@{}rl@{\extracolsep{\fill}}rrr|rrr@{}}
\toprule
& & \multicolumn{3}{c|}{\textbf{2025 A-board}} & \multicolumn{3}{c}{\textbf{2026 B-board (final)}} \\
\textbf{Rank} & \textbf{Track 2 system} & SR & AR\% & MDD\% & SR & AR\% & MDD\% \\
\midrule
1 & FinArbi & 1.56 & 28.18 & -11.93 & \textbf{1.97} & 32.81 & -8.55 \\
2 & lizheng & 1.70 & 26.26 & -10.38 & 1.71 & \textbf{34.34} & -8.12 \\
3 & FinanceAgent & 1.58 & 33.99 & -15.10 & 1.27 & 12.15 & -11.26 \\
4 & AEGIS & 1.87 & 23.12 & -6.72 & 0.85 & 7.17 & \textbf{-3.00} \\
5 & Turing & 1.42 & 45.25 & -22.02 & 0.79 & 23.49 & -17.57 \\
6 & AlphaAllocator & 1.60 & 32.15 & -11.09 & 0.76 & 15.32 & -7.49 \\
7 & Grilled Egg Crepe & 1.40 & 32.13 & -18.42 & 0.71 & 14.57 & -10.68 \\
8 & AgentWeaver & 1.55 & 26.29 & -10.68 & 0.70 & 6.44 & -5.81 \\
9 & Texas Hold'em & 0.95 & 15.35 & -12.27 & 0.66 & 11.50 & -9.45 \\
10 & Varina777 & \textbf{2.01} & 33.63 & -9.89 & 0.62 & 9.05 & -11.92 \\
11 & Uibequant\_strategy & 1.98 & \textbf{45.55} & -9.43 & 0.21 & 2.19 & -14.72 \\
12 & Shrike & 1.59 & 36.63 & -16.91 & -0.16 & -4.62 & -12.16 \\
13 & Dianjin & 1.34 & 14.08 & \textbf{-6.46} & -1.30 & -13.03 & -7.65 \\
\bottomrule
\end{tabular*}
\caption{Official Track~2 results, ordered by final B-board Sharpe ratio.}
\label{tab:track2-results}
\end{table}

\subsection{Cross-System Findings}

\paragraph{LLM capability and investment performance.}
General-purpose capability does not directly translate into investment capability. Under the fixed CPA workflow, the leading LLM varies across the reported settings, and GPT-5, Gemini-2.5-Pro, DeepSeek, and Qwen models exhibit different return and risk profiles across the two allocation tasks. Investment performance reflects the interaction of LLM behavior with signal selection, calibration, trading cadence, and risk control.

\paragraph{Hybrid architectures.}
Most submitted systems use hybrid LLM--quantitative architectures. Language models extract events, sentiment, regimes, confidence scores, or bounded asset views; quantitative and deterministic modules perform sizing, diversification, turnover control, and execution validation. Independently developed systems also converge on rolling memory, delayed confirmation, minimum holding periods, and fallback rules. These mechanisms suppress isolated headlines and excessive turnover. Longer signal persistence can delay responses to new information, creating a trade-off between responsiveness and stability.

\paragraph{Investment Agent harness.}
FinArbi, FinanceAgent, and AgentWeaver use the same LLM, DeepSeek-V4-Flash, with different signal-arbitration, memory, allocation, and safeguard mechanisms, and obtain different B-board results. AlphaAllocator combines local representation and sentiment models with a constrained quantitative pipeline and ranks first in Track~1. The \emph{investment agent harness} comprises the observation, orchestration, memory, calibration, portfolio-construction, and safety mechanisms surrounding an LLM. Financial agent evaluation consequently covers the complete LLM--harness system.

\section{Conclusion and Discussion}

This work proposes CN‑Buzz2Portfolio, the official evaluation benchmark for NLPCC 2026 Shared Task 4, providing matched financial news‑ETF data, a standard backtesting framework, and a reproducible baseline. Evaluations over the baseline and 14 submitted systems yield three key observations: stronger LLM reasoning does not guarantee higher investment returns; hybrid LLM‑quant pipelines combining text semantics, quantitative features and risk constraints achieve better performance; auxiliary agent modules exert larger influence on portfolio outcomes than base LLMs. Our open‑sourced resources lay a foundation for future research on financial investment agents.

All results are for academic purposes only and do not constitute investment advice. This benchmark still suffers from limited interpretability and practical transferability, while uncontrolled component adjustments obscure causal attribution of performance differences. Future work will adopt controlled ablation studies, enrich evaluation metrics, introduce diverse financial data and realistic trading frictions to narrow the gap between simulation and real‑world  trading.

\end{document}